\documentclass[conference]{IEEEtran}
\IEEEoverridecommandlockouts
\usepackage{cite}
\usepackage{amsmath,amssymb,amsfonts}
\usepackage{algorithmic}
\usepackage{graphicx}
\usepackage{textcomp}
\usepackage{xcolor}
\usepackage{array}
\usepackage{multirow}
\usepackage{enumerate}
\usepackage{url}
\usepackage{arydshln}
\usepackage{multirow}
\usepackage{booktabs}

\def\BibTeX{{\rm B\kern-.05em{\sc i\kern-.025em b}\kern-.08em
    t\kern-.1667em\lower.7ex\hbox{E}\kern-.125emX}}

\usepackage{fancyhdr}
\begin{document}

\title{Using Knowledge-Embedded Attention to Augment Pre-trained Language Models for Fine-Grained Emotion Recognition 
\thanks{This research is supported in part by the National Research Foundation, Singapore under its AI Singapore Program (AISG Award No: AISG2-RP-2020-016), and a Singapore Ministry of Education Academic Research Fund Tier 1 grant to DCO.}
} 

\author{


\IEEEauthorblockN{Varsha~Suresh\textsuperscript{1}, Desmond~C.~Ong\textsuperscript{2,3}}
\IEEEauthorblockA{
\textit{\textsuperscript{1}Department of Computer Science, National University of Singapore} \\
\textit{\textsuperscript{2}Department of Information Systems and Analytics, National University of Singapore} \\
\textit{\textsuperscript{3}Institute of High Performance Computing, Agency for Science, Technology and Research, Singapore} \\
varshasuresh@u.nus.edu,
dco@comp.nus.edu.sg}

}

\maketitle
\thispagestyle{fancy}

\begin{abstract}

Modern emotion recognition systems are trained to recognize only a small set of emotions, and hence fail to capture the broad spectrum of emotions people experience and express in daily life. In order to engage in more empathetic interactions, future AI has to perform \textit{fine-grained} emotion recognition, distinguishing between many more varied emotions. Here, we focus on improving fine-grained emotion recognition by introducing external knowledge into a pre-trained self-attention model. We propose Knowledge-Embedded Attention (KEA) to use knowledge from emotion lexicons to augment the contextual representations from pre-trained ELECTRA and BERT models. Our results and error analyses outperform previous models on several datasets, and is better able to differentiate closely-confusable emotions, such as afraid and terrified.
\end{abstract}


\section{Introduction}

Imagine telling your chatbot that your dog just died. Instead of correctly understanding that you are experiencing grief (and offering condolences), it classifies you as feeling sad and offers to play you a happy song to cheer you up. People experience a wide range of emotions, and it is important for AI agents to correctly recognize subtle differences between emotions like sadness and grief, in order to improve their interactions with people and to avoid making a \emph{faux pas} like the chatbot above \cite{ huang2020challenges}. Traditionally, the vast majority of work in emotion recognition from text focuses on recognizing just six ``basic'' emotions \cite{poria2019emotion,alswaidan2020survey}, usually happiness, surprise, sadness, anger, disgust, and fear. This set clearly fails to capture the broad spectrum of emotions that people experience and express in daily life, such as pride, guilt, and hope \cite{Skerry15neuralrepresentations, cowen2017self,scherer2013human}.

Recently, there have been efforts to focus on larger classes of emotions from text-based data with the introduction of the EmpatheticDialogues dataset \cite{rashkin-etal-2019-towards}, which consists of online conversations in 32 different emotion categories, and the GoEmotions dataset \cite{demszky-etal-2020-goemotions}, which consists of Reddit comments labelled with 28 different classes. These recently-proposed datasets are an important step in training fine-grained emotion classification models that can recognize more nuanced emotions. 

Concurrently, pre-trained language models such as ELECTRA \cite{clark2020electra} and BERT \cite{devlin-etal-2019-bert} have achieved state-of-the-art performance in NLP, such as on various text-classification tasks. Moreover, incorporating knowledge into text representations have been shown to improve model performance in various domains \cite{roy2020incorporating,shin2017lexicon}. Indeed, much of the differences between fine-grained emotion classes require deeper semantic knowledge, which may already exist in resources like emotion lexicons. Borrowing from these insights, we hypothesized that incorporating such external knowledge into existing contextualized representations will improve model performance for fine-grained emotion recognition.

In this work, we introduce Knowledge-Embedded Attention (KEA), a knowledge-augmented attention mechanism that enriches the contextual representation provided by pre-trained language models using emotional information obtained from external knowledge sources. This is achieved by incorporating the encoded emotional knowledge with the contextual representations to form a modified key matrix. This key matrix is then used to attend to the contextual representations to construct a more emotionally-aware representation of the input text that can be used to recognise emotions. We introduce two variants of KEA, (i) a word-level KEA and (ii) a sentence-level KEA, which incorporate knowledge at different text granularities.

We compare our approach with representative baselines and find that KEA-based models show improved performance for fine-grained emotion recognition. Furthermore, we perform additional analysis to show the extension of our model to generalise to other contextual encoders and also shows its efficacy using other emotion knowledge sources. Finally, we perform an in-depth case study using the EmpatheticDialogues dataset to look into the two categories of emotion classes that contribute to a majority of misclassifications, to investigate the impact of using KEA on both these categories.

\section{Related Work}

\subsection{Emotion recognition from text}

In recent years, emotion recognition systems are primarily modelled using neural architectures such as LSTMs, RNNs and CNNs \cite{su2018lstm, colneric2020emotion, alswaidan2020survey, wu2019attending} as they tend to outperform classical machine learning approaches that use feature-engineering \cite{alswaidan2020survey}. The most recent pre-trained language models use Transformer-based architectures such as ELECTRA \cite{clark2020electra} and BERT \cite{devlin-etal-2019-bert}, and have achieved state-of-the-art performance in a variety of downstream tasks in NLP. These pre-trained language models have also been employed for emotion recognition \cite{huang2019emotionx, demszky-etal-2020-goemotions, rashkin-etal-2019-towards}. However, the vast majority of the aforementioned approaches often only consider a small set of 6-8 emotion classes \cite{poria2019emotion}, such as happiness, surprise, sadness, anger, disgust, and fear. 

\subsection{Fine-grained emotion recognition}

Many AI papers implicitly rely on psychological theories that emotions exist as discrete categories, and thus formulate emotion recognition as a classification problem.
Many researchers borrow Ekman's \cite{ekman1999basic} list of six ``basic" emotions, or similar lists like Plutchik's \cite{robert2001the} eight ``primary" emotions. However, these lists are far from comprehensive. People in their daily lives obviously experience a much larger set of emotions, including shame, guilt, and pride, which the majority of emotion recognition models today, being trained on a limited set of emotions, will fail to capture. 


We use the term ``fine-grained" emotion classification to indicate tasks with a larger number of emotion classes (minimally, greater than 8). Fine-grained emotion recognition systems are gaining traction due to their importance in the development of empathetic agents that can differentiate subtle and complex emotions. 
The major limiting factor has been the lack of datasets with fine-grained emotion labels. 
%
This has changed with the introduction of recent corpora such as EmpatheticDialogues \cite{rashkin-etal-2019-towards}, which consists of textual conversations labelled with 32 emotions, and GoEmotions \cite{demszky-etal-2020-goemotions}, which consists of Reddit comments labelled with 28 emotions. In this work, we enhance contextual embeddings from pre-trained models such as ELECTRA using lexicon knowledge to build an emotion recognition model that scales well to fine-grained emotions.   


\subsection{Knowledge-enhanced text representations}

External knowledge sources are known to provide explicit knowledge for the task at hand, which can complement the representations implicitly learnt by deep learning models \cite{roy2020incorporating}. Knowledge has been incorporated with representations learnt by neural architectures such as BiLSTMs, CNNs and RNNs, using different techniques such as concatenation \cite{de2019joint} and attention-based mechanisms \cite{margatina-etal-2019-attention,ma2018sentic,shin2017lexicon}. However, given the improvements offered by pre-trained language models like BERT, we focus the rest of this discussion on works which incorporate knowledge into these pre-trained models to further enhance their performance.

Integrating knowledge into contextual representation from pre-trained language models can be classified into two approaches. The first type of approach re-trains these language models from scratch by modifying the raw input \cite{poerner-etal-2020-e,zhang-etal-2019-ernie}, via multi-task learning \cite{levine-etal-2020-sensebert,tian-etal-2020-skep} or by augmenting the word-embedding \cite{zhong2019knowledge,ke-etal-2020-sentilare,chen2020kgpt,liu2021kg}. Although, these methods can help improve performance in downstream tasks, they have to be retrained each time and/or redesigned if other knowledge sources need to be accommodated.

The second set of approaches enrich the contextual representation broadly either via early fusion or late fusion. Early fusion techniques incorporate knowledge source at the input/embedding stage. Generally the knowledge sources used in early fusion are linguistic in nature such as auxiliary sentences \cite{wu2020contextguided}. Late fusion, on the other hand, involves combining the knowledge embedding of the input and the contextual representation at the later stage. Numerous approaches exist to do this. A common way of performing late fusion is by concatenating the external knowledge embedding and contextual representations \cite{babanejad-etal-2020-affective,ostendorff2019enriching}. Alternatively, Bruyne et al., \cite{de2019joint} combined BERT representation with lexicon data using word-level concatenation and passed it via a BiLSTM to perform classification. Additionally,
Wang et al.,\cite{wang2020kadapter} combined knowledge sources by pre-training knowledge data separately using models called adapters and combined them with BERT representations at a later stage.  These set of approaches provide flexibility to add additional knowledge sources without the need to modify or re-train the language model depending on the kind of knowledge added.


We also consider emotional knowledge provided by lexicon data, which largely consists of association scores for emotional dimensions such as emotion intensity, valence and arousal. These ratings are generally just real-valued vectors \cite{mohammad2018word, mohammad-2018-obtaining}. 
Hence, we focus on incorporating emotional knowledge to the contextual embeddings produced by pre-trained language models via late fusion and fine-tuning them to aid in the task of emotion recognition. 
\begin{figure*}[t]
    \centering
    \includegraphics[width=0.65\linewidth]{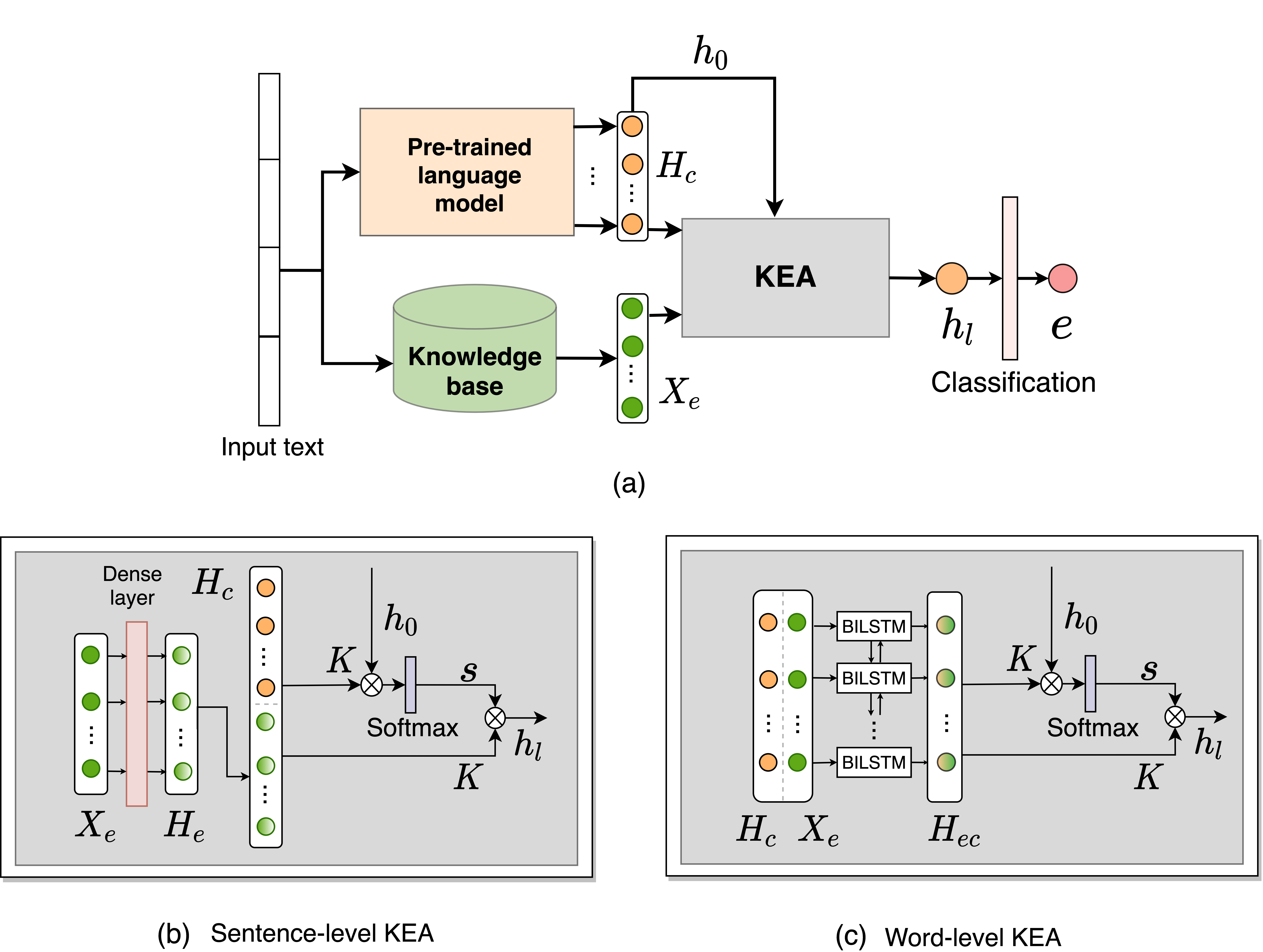}
  \caption{An overview of the proposed KEA approach. (a) The overall flow of KEA-based models. Here, $X_{e}$ is the transformed input text using the Knowledge base and $H_{c}$ is the last layer output of the pre-trained language model (e.g., ELECTRA). (b) Sentence-level KEA, where $H_{e}$ denotes the emotional encoding that is concatenated with $H_{c}$ to form the Key matrix of attention. (c) Word-level variant of KEA where $H_{c}$ and $X_{e}$ are concatenated at the word-level and passed into a BiLSTM to obtain $H_{ec}$ which serves as the Key matrix for attention.}
  \label{fig:model} 
\end{figure*}

\section{Proposed Approach}

\subsection{Model Description}

The proposed model, in Fig. \ref{fig:model}, embeds an input text $X$ into two latent spaces: a (i) contextual representation, and an (ii) emotional encoding. The emotional encoding is obtained from external knowledge sources such as emotional lexicons. We hypothesize that enriching contextual-representation using emotional encodings via Knowledge-Embedded Attention (KEA) will provide a richer representation of the input text, improving the final emotion classification. We introduce both a sentence-level and a word-level variant of KEA.

\textbf{Contextual representations} provided by pre-trained language models have been shown to improve language understanding by paying attention to all words and their surrounding context to encode a meaningful representation of the content \cite{devlin-etal-2019-bert, clark2020electra}. In our work we use both BERT and the recently-introduced ELECTRA, a discriminatively pre-trained Transformer model which achieves state-of-the-art performance in various downstream tasks.

We denote the representation corresponding to the hidden states from the last layer of the pre-trained models as matrix $H_{c} = \{h_0, \cdots, h_{N}\}$ where $H_{c} \in \mathbb{R}^{N \times l_{c}}$ and $l_{c}$ is the size of the output representation generated by BERT or ELECTRA (we use the \texttt{-base} versions of BERT and ELECTRA; $l_{c}$=768). The representation corresponding to the [CLS] token, $h_{0}$, is taken as the contextual representation for the entire input text for classification. Intuitively, this offers a meaningful summary of the entire input which is further enriched with external emotional knowledge via KEA.

\textbf{Knowledge-Embedded Attention (KEA)} incorporates emotional knowledge obtained from lexicons with contextual representations provided by pre-trained language models. To achieve this, we use knowledge obtained from emotion lexicons, which provides ratings associated to different emotion-related dimensions. Below we elaborate on the two proposed variants of KEA.

\noindent\textbf{Sentence-level:} In sentence-level KEA (Fig. \ref{fig:model}a), we obtain the emotional encoding by transforming the input $X$ to a feature vector $X_{e}$, whose dimension depends on the lexicon data used, which we denote as $l_{e}$. The input sequence is also padded to a dimension of fixed value, which we set to 512 as it is the default sequence length dimension of \texttt{-base}. We then project the above transformed input vectors using dense layers to form the sentence-level emotional encoding $H_{e}$ where $H_{e} \in \mathbb{R}^{l_{e} \times l_{c}}$.

Following self-attention terminology \cite{vaswani2017attention}, we concatenate the emotional encoding $H_{e}$ and contextual representation $H_{c}$ to form the Key $K$ where $K \in \mathbb{R}^{(N+l_{e})\times l_{c}}$ and we use $h_{0}$ as the Query to obtain the softmax-attention score $s$. Intuitively, $H_{c}$ provides a meaningful representation of the entire input based on its pre-trained knowledge, and $H_{e}$ provides an emotional summary of the input based on the lexicon information. The final representation $h_{l}$ is obtained by weighting the key matrix with $s$. Including $H_{c}$ together with $H_{e}$ into the key $K$, helps preserve the contextual information learnt by the encoder in addition to the added emotional knowledge while re-weighting $h_{0}$. The overall attention layer is given by:
\begin{align}
     K = \text{concat}(H_{c},H_{e}) \qquad \nonumber 
     s &= \text{softmax}(h_{0}^\intercal \cdot K)\\
     h_{l} &= s^\intercal \cdot K \label{eq:kea_sen}
\end{align}

\noindent{\textbf{Word-level}}: In word-level KEA (Fig. \ref{fig:model}b) we modify $K$ by incorporating knowledge at the word-level. We transform each word $x_{i}$ in the input $X$ to a feature vector $x^{e}_{i}$. Each contextual representation $h_{i}$ (i.e., from ELECTRA) is concatenated with the corresponding knowledge information $ x^{e}_{i}$, and is then projected into a latent state using a BiLSTM. 
We denote the hidden output state generated by the BiLSTM as $h^{ec}_{i}$:
\begin{equation}
\small{h^{ec}_{i} = \text{BiLSTM}([h_{i}; x^{e}_{i}],h^{ec}_{i-1},h^{ec}_{i+1}), \forall i \in [1, . . . ,N]}    
\end{equation}
We use $H_{ec} = (h^{ec}_{0}, \cdots h^{ec}_{N})$ where $H_{ec} \in \mathbb{R}^{N \times 2l}$, where $l$ is hidden state dimension of the LSTM which we set to 384. In word-level KEA, $H_{ec}$ serves as the Key matrix $K$. The remaining steps follows sentence-level KEA:
\begin{align}
     K = \text{concat}(H_{ec}) \qquad \nonumber 
     s &= \text{softmax}(h_{0}^\intercal \cdot K)\\
     h_{l} &= s^\intercal \cdot K \label{eq:kea_word}
\end{align}
\textbf{Classification:} Finally, $h_{l}$ is fed into a two-layer dense network to get the output probabilities of the emotions for the corresponding input. We train the model using the standard Cross Entropy loss for single-label settings and Sigmoid cross entropy loss for multi-label settings \cite{demszky-etal-2020-goemotions}.
\section{Evaluation}
\subsection{Datasets}
We tested our models on three datasets which span a range of fine-grained emotion classes (11, 28, 32 classes) and text domains (tweets, forum posts, and conversations).
\begin{itemize}
\item \textbf{EmpatheticDialogues (ED)} \cite{rashkin-etal-2019-towards}\footnote{\url{https://github.com/facebookresearch/EmpatheticDialogues}} This dataset consists of 24,850 two-way conversations in English with an average of 4.31 utterances. Each conversation is annotated with one of 32 emotion labels and the label distribution of the dataset is balanced. Note that the labels are for the entire conversation and not for each utterance. The train/validation/test split for the dataset is 19,533 / 2,770 / 2,547 samples respectively. For our input, we concatenate the utterances separating them with the [SEP] token. 
\item\textbf{GoEmotions} \cite{demszky-etal-2020-goemotions}\footnote{\url{https://github.com/google-research/google-research/tree/master/goemotions}} 
The (filtered) version of this dataset comprises 54k English Reddit comments annotated with one or more of 28 classes. The train/validation/test split is 43,410 / 5,426 / 5,427 samples respectively.
\item \textbf{Affect in Tweets (AIT)} \cite{mohammad-kiritchenko-2018-understanding} This dataset was part of SemEval-2018 Task 1: Affect in Tweets\footnote{\url{https://competitions.codalab.org/competitions/17751}} 
and consists of Twitter data. We utilise the dataset provided for the E-c task where each tweet is classified as one, or more, of 11 emotional states of the tweeter. 
The dataset comprises a total of 10,983 tweets with the train/validation/test split as 6,838 / 886 / 3,259 samples respectively.   
\end{itemize}
\subsection{Evaluation Metrics}
For single-label settings (EmpatheticDialogues dataset) we use top-1 accuracy, top-3 accuracy (henceforth referred to as top-1 and top-3 respectively) and macro-F1 score. For multi-label settings, we use macro F1-score, Precision and Recall. In addition, for Affect in Tweets we also report the Jaccard index, which was the primary evaluation metric in SemEval-2018 E-c task.
\subsection{Lexicon features}
All results in Table \ref{tab:main_res} were obtained using NRC-VAD \cite{mohammad-2018-obtaining} (henceforth referred to as VAD) lexicon data. This knowledge source contains ratings of valence, arousal, and dominance of 20k words. The rating values range from 0 to 1 and vary from negative to positive (valence), calm to aroused (arousal), and submissive to dominant (dominance). To transform an input text $X$ into its corresponding valence, arousal, and dominance vectors, we replace each word in the utterances with the corresponding values from the lexicon. Following previous works \cite{zhong2019knowledge}, words which do not appear in the lexicon are given the mid-value score of 0.5. The choice of knowledge can be varied based on the task at hand; we show the efficacy of KEA-based methods by incorporating another knowledge source in Section \ref{sec:other_knowledge}. 
\subsection{Baselines}
\label{sec:baselines}
We compare the performance of our model with three main categories of baselines.
\begin{itemize}
    \item \textbf{Models without pre-training}: We compare with recurrent models such as (i) BiLSTM with self-attention \cite{lin2017structured}, (ii) CNN+c-LSTM \cite{poria-etal-2017-context} and (iii) RCNN \cite{lai2015recurrent}.
    \item \textbf{Pre-trained language models}: ELECTRA \cite{clark2020electra} and BERT \cite{devlin-etal-2019-bert}. We obtain the pre-trained models \texttt{bert-base-uncased} and \texttt{electra-base-discriminator} from HuggingFace’s Transformers library \footnote{\url{https://huggingface.co/transformers/}}.
    \item \textbf{Knowledge-enhanced models}: We compare with commonly used methods of incorporating knowledge with contextual representations: 
    
    (i) $\text{k-ELECTRA}_{\text{concat}}$, which is a simple concatenation emotional encoding obtained from $H_{e}$ to $H_{c}$, which is then projected using dense layers to perform classification. This is the most straightforward way to incorporate knowledge \cite{ostendorff2019enriching,babanejad-etal-2020-affective}. This serves as baseline for sentence-level knowledge incorporation. By contrast, our KEA includes an attention layer. 
    
    (ii) $\text{k-ELECTRA}_{\text{BiLSTM}}$ which is similar to word-level KEA, concatenates knowledge in a word-level fashion to the contextual representation. This representation is further passed via single-layer BiLSTM similar to \cite{de2019joint}, with the hidden state dimension of the BiLSTM set to 384. This serves as a baseline to compare word-level knowledge incorporation. 
    
    (iii) KET (Knowledge Enriched Transformers) \cite{zhong2019knowledge}, a knowledge-based dynamic graph attention model that enhances Transformers using VAD and ConceptNet \cite{speer2017conceptnet} to detect emotion from conversation data.

    \item In addition to the above models, we compared with state-of-the-art (SOTA) performance for all the datasets. For ED we choose Attention Gated Hierarchical Memory Network (AGHMN) \cite{jiao2020real} that uses HMN and GRU based hierarchical architecture to capture utterance-level emotions from conversation, for Affect in Tweets we compared with performance of the best team in the SemEval-2018 E-c challenge \cite{baziotis2018ntua} and for GoEmotions we compared with performance provided by authors of the dataset \cite{demszky-etal-2020-goemotions}. 
\end{itemize}
As AGHMN and KET is designed primarily for conversation and require labels for each utterance, we report the performance only on ED dataset and we label every utterance in a conversation in the ED dataset with the conversation label to make these models compatible with it. 

\begin{table*}[t]
\centering
\resizebox{\linewidth}{!}{
\begin{tabular}{lccc|cccc|ccc}
\hline
 &
  \multicolumn{3}{c}{\textbf{EmpatheticDialogues}} &
  \multicolumn{4}{c}{\textbf{Affect in Tweets}} &
  \multicolumn{3}{c}{\textbf{GoEmotions}} \\ \hline
 \multicolumn{1}{l|}{} &
  \textbf{\begin{tabular}[c]{@{}c@{}}top-1 / \%\end{tabular}} &
  \textbf{\begin{tabular}[c]{@{}c@{}}top-3 / \%\end{tabular}} &
  \multicolumn{1}{c|}{\textbf{F1}} &
  \textbf{jaccard} &
  \textbf{precision} &
  \textbf{recall} &
  \multicolumn{1}{c|}{\textbf{F1}} &
  \textbf{precision} &
  \textbf{recall} &
  \textbf{F1} \\ \cline{2-11} 
\multicolumn{1}{l|}{BiLSTM \cite{lin2017structured}}&
  35.8 (0.6) &
  62.0 (0.8) &
  35.9 (0.6) &
  43.7 (1.0) &
  41.8 (1.9) &
  54.7 (2.2) &
  46.7 (0.4) &
  56.5 (2.6) &
  39.3 (2.3) &
  43.9 (1.0) \\
  		
\multicolumn{1}{l|}{c-LSTM \cite{poria-etal-2017-context}} &
  37.9 (0.1) &
  64.7 (0.3) &
  37.5 (0.2) &
  51.8 (0.5) &
  45.3 (0.8) &
  62.4 (0.6) &
  51.0 (0.8) &
  50.9 (1.7) &
  27.2 (0.8) &
  31.6 (1.0) \\
\multicolumn{1}{l|}{RCNN \cite{lai2015recurrent}} &
  43.0 (0.6) &
  69.5 (0.4) &
  43.2 (0.6) &
  54.2 (0.5) &
  46.7 (1.4) &
  64.1 (1.9) &
  53.5 (0.5) &
  \textbf{58.4 (1.0)} &
  37.5 (1.0) &
  42.5 (0.6) \\
\multicolumn{1}{l|}{SOTA}&
  - &
 -&
  $\text{41.2}^{1}$ (-)&
  $\text{57.8}^{2}$ (-) &
  - &
  - &
  - &
  $\text{40.0}^{3}$ (-)&
  $\text{63.0}^{3}$ (-)&
  $\text{46.0}^{3} (-)$ \\ \hline
\multicolumn{1}{l|}{BERT} &
  51.9 (0.6) &
  78.2 (0.5) &
  50.7 (1.0) &
  56.3 (0.8) &
  54.2 (2.6) &
  64.1 (3.8) &
  57.7 (0.4) &
  51.7 (1.9) &
  49.5 (2.3) &
  48.3 (1.5) \\
\multicolumn{1}{l|}{ELECTRA}&
  52.8 (0.5) &
  78.7 (0.4) &
  50.9 (0.7) &
  57.6 (0.2) &
  57.2 (1.7) &
  61.2 (1.9) &
  57.6 (1.2) &
  47.4 (1.3) &
  50.4 (1.7) &
  47.5 (0.7) \\ \hline
  \multicolumn{1}{l|}{KET \cite{zhong2019knowledge}} &
   36.2 (-)&
   - &
   34.9 (-)&
  - &
  - &
  - &
  - &
  - &
  - &
  - \\
\multicolumn{1}{l|}{$\text{k-ELECTRA}_{\text{bilstm}}$} &
  48.1 (0.9) &
  75.0 (1.5) &
  45.6 (1.0) &
  54.9 (1.3) &
  39.7 (2.6) &
  \textbf{68.2 (2.5)} &
  49.5 (2.2) &
  43.8 (2.7) &
  44.8 (1.7) &
  42.3 (0.9) \\
\multicolumn{1}{l|}{$\text{k-ELECTRA}_{\text{concat}}$} &
  52.1 (0.4) &
  78.0 (0.5) &
  50.3 (0.7) &
  55.7 (1.8) &
  47.3 (3.2)&
  66.3 (1.7) &
  54.3 (2.6)&
  45.7 (1.4) &
  48.2 (0.9) &
  45.6 (0.9) \\
\multicolumn{1}{l|}{$\text{KEA-ELECTRA}_{\text{word}}$} &
  53.6 (0.6) &
  78.5 (0.8) &
  52.5 (0.6) &
  57.7 (0.8) &
  50.8 (0.8) &
  66.9 (1.1) &
  57.1 (0.6) &
  46.1 (1.8) &
  50.2 (0.9) &
  46.8 (0.7) \\
\multicolumn{1}{l|}{$\text{KEA-ELECTRA}_{\text{sentence}}$} &
  \textbf{54.1 (0.6)} &
  \textbf{80.5 (0.5)} &
  \textbf{53.1 (0.7)} &
  \textbf{58.3 (0.1)} &
  \textbf{57.7 (1.4)} &
  61.9 (0.7) &
  \textbf{59.1 (0.3)} &
  48.6 (0.9) &
  \textbf{52.9 (0.6)} &
  \textbf{49.6 (0.8)} \\ \hline \\
\end{tabular}
}
\caption{Summary of the results obtained using test data for EmpatheticDialogues (ED), Affect in Tweets (AIT) and GoEmotions datasets. SOTA row implies the state-of-the-art performance on the three datasets, 1 is obtained by using  AGHMN\cite{jiao2020real} model which predicts emotion from textual conversations, 2 and 3 are taken from \cite{baziotis2018ntua} and \cite{demszky-etal-2020-goemotions} respectively.  For ED, we use top-1 accuracy, top-3 accuracy and macro-F1. KET and AGHMN are designed for conversation, hence we consider their performance only for ED. For the AIT and GoEmotions datasets, we compare the performances using Precision, Recall and macro-F1. For AIT we also report the Jaccard index. The metrics are averaged over 5 runs, with standard deviations reported in parentheses and best scores in bold.}
\label{tab:main_res}
\vspace{-3mm}
\end{table*}

\subsection{Implementation details}
Input text was converted into tokens using WordPiece tokenization followed by ELECTRA preprocessing. For fine-tuning, we use Adam optimizer \cite{kingma2015Adam} and each input text in the batch is padded to the length of the text with the maximum length. We repeated this process with five random seeds and report the mean and the standard deviation of performance over 5 runs. For running our models, we used a Google Colaboratory instance equipped with NVIDIA Tesla T4 GPU.

For fine-tuning KEA-based models, we chose learning rates from the set \{$1 \mathrm{e}{-05}$, $2 \mathrm{e}{-05}$, $3 \mathrm{e}{-05}$\} and batch size from the set \{10,16\}. We used the Adam \cite{kingma2015Adam} optimiser with $\beta_{1}$ set to 0.9, $\beta_{2}$ set to 0.999, and $\epsilon$ set to 1e-08. Early stopping was done based on top-1 accuracy in the validation set for EmpatheticDialogues dataset and F1 score in Affect in Tweets and GoEmotions dataset. For Affect in Tweets dataset the input tweets were prepossessed by removing elements such as non-ascii characters, letter repetitions and extra white-spaces and replacing all the user-mentions and links to unique identifiers. We provide source code for all the implementations\footnote{\url{https://github.com/varsha33/Fine-Grained-Emotion-Recognition}}. 

For the BiLSTM, cLSTM and RCNN models, we used pre-trained GloVe vectors of dimension 200 as word embedding. These models were trained with a batch size of 64 using Adam optimiser and learning rate was chosen from the set \{$1 \mathrm{e}{-02}$, $1 \mathrm{e}{-03}$, $5 \mathrm{e}{-03}$\} based on the option that yielded the best top-1 accuracy in the validation set for EmpatheticDialogues dataset, and F1 score in Affect in Tweets and GoEmotions dataset. For the comparison with KET\footnote{\url{https://github.com/zhongpeixiang/KET}} and AGHMN\footnote{\url{https://github.com/wxjiao/AGHMN}} we used the implementation provided by the authors. For fair comparison we only compared with these methods for the EmpatheticDialogues dataset by applying the conversation-level label to every utterance in the conversation as these methods perform utterance-level emotion recognition taking into account the sequential nature of the conversation. 

\section{Results and Discussion}

\begin{table*}[ht!]
\centering
\resizebox{\linewidth}{!}{
\begin{tabular}{lcccccccccc}
\hline
 &
  \multicolumn{3}{c}{\textbf{EmpatheticDialogues}} &
  \multicolumn{4}{c}{\textbf{Affect in Tweets}} &
  \multicolumn{3}{c}{\textbf{GoEmotions}} \\ \hline
\multicolumn{1}{l|}{} &
  \textbf{\begin{tabular}[c]{@{}c@{}}top-1 / \%\end{tabular}} &
  \textbf{\begin{tabular}[c]{@{}c@{}}top-3 / \%\end{tabular}} &
  \multicolumn{1}{c|}{\textbf{F1}} &
  \textbf{jaccard} &
  \textbf{precision} &
  \textbf{recall} &
  \multicolumn{1}{c|}{\textbf{F1}} &
  \textbf{precision} &
  \textbf{recall} &
  \textbf{F1} \\ \cline{2-11} 
\multicolumn{1}{l|}{BERT}&
  51.9 (0.6) &
  78.2 (0.5) &
  \multicolumn{1}{c|}{50.7 (1.0)} &
  56.3 (0.8) &
  54.2 (2.6) &
  64.1 (3.8) &
  \multicolumn{1}{c|}{57.7 (0.4)} &
  \textbf{51.7 (1.9)} &
  49.5 (2.3) &
  48.3 (1.5) \\\hline
\multicolumn{1}{l|}{$\text{KEA-BERT}_{\text{word}}$} &
  51.8 (0.3) &
  77.5 (0.4) &
  \multicolumn{1}{c|}{51.0 (0.3)}&
  56.9 (0.3) &
  51.9 (1.3) &
  \textbf{66.1 (1.2)} &
  \multicolumn{1}{c|}{57.7 (0.5)} &
  46.2 (0.8) &
  51.1 (0.8) &
  47.2 (0.7) \\
\multicolumn{1}{l|}{KEA-BERT$_{\text{sentence}}$} &
  \textbf{53.3 (0.4)} &
  \textbf{79.3 (0.7)} &
  \multicolumn{1}{c|}{\textbf{52.4 (0.5)}} &
  \textbf{57.0(0.6)} &
  \textbf{56.8 (1.5)} &
  61.3 (1.4) &
  \multicolumn{1}{c|}{\textbf{58.2 (0.2)}} &
  51.4 (2.2) &
  \textbf{52.5 (0.6)} &
  \textbf{51.0 (0.7)} \\ \hline \\

\end{tabular}
}
\caption{Generalization to other contextual encoders: Summary of the results obtained using test data, for another contextual encoder, BERT. The metrics are averaged over 5 runs, with standard deviations reported in parentheses and best scores in bold.}
\label{tab:context}
\end{table*}

\begin{table*}[ht!]
\centering
\resizebox{\linewidth}{!}{
\begin{tabular}{llccc|cccc|ccc}
\hline
 &
   &
  \multicolumn{3}{c}{\textbf{EmpatheticDialogues}} &
  \multicolumn{4}{c}{\textbf{Affect in Tweets}} &
  \multicolumn{3}{c}{\textbf{GoEmotions}} \\ \hline
\multicolumn{1}{l|}{} &
  \multicolumn{1}{l|}{\textbf{Lexicon}} &
  \textbf{top-1 / \%} &
  \textbf{top-3 / \%} &
  \textbf{F1} &
  \textbf{jaccard} &
  \textbf{precision} &
  \textbf{recall} &
  \textbf{F1} &
  \textbf{precision} &
  \textbf{recall} &
  \textbf{F1} \\ \cline{2-12} 

\multicolumn{1}{l|}{ELECTRA} &
  - &
  \multicolumn{1}{|l}{52.8 (0.5)} &
  78.7 (0.4) &
  50.9 (0.7) &
  57.6 (0.2) &
  \textbf{57.2 (1.7)} &
  61.2 (1.9) &
  57.6 (1.2) &
  47.4 (1.3) &
  50.4 (1.7) &
  47.5 (0.7) \\
\multicolumn{1}{l|}{$\text{k-ELECTRA}_{\text{concat}}$} &
  EIL &
  \multicolumn{1}{|l}{51.7 (0.6)}&
  77.7 (0.9) &
  50.1 (0.8) &
  55.7 (1.3) &
  46.1 (3.4) &
  60.9 (4.4) &
  51.4 (2.3) &
  46.4 (0.8) &
  47.5 (1.0) &
  45.0 (0.7) \\ \hline
\multicolumn{1}{l|}{$\text{KEA-ELECTRA}_{\text{word}}$} &
  EIL &
  \multicolumn{1}{|l}{53.7 (0.6)} &
  79.2 (0.8) &
  52.7 (0.7) &
  57.7 (0.8) &
  53.3 (3.6) &
  65.1 (2.5) &
  57.1 (1.7) &
  44.9 (2.0) &
  50.8 (1.2) &
  46.5 (0.6) \\
\multicolumn{1}{l|}{$\text{KEA-ELECTRA}_{\text{sentence}}$} &
  EIL &
  \multicolumn{1}{|l}{\textbf{54.0 (0.5)}} &
  \textbf{80.1 (0.8)} &
  \textbf{53.0 (0.7)} &
   \textbf{58.1 (0.4)} &
   52.8 (2.1) &
  \textbf{66.5 (2.6)} &
  \textbf{58.1 (0.2)} &
  \textbf{47.5 (0.8)} &
  \textbf{53.0 (0.5)} &
  \textbf{49.2 (0.8)} \\ 
  \hline \\
\end{tabular}
}
\caption{Generalization to other lexicons: Summary of the results obtained using test data, while using NRC-Emotion Intensity Lexicon (EIL). The metrics are averaged over 5 runs, with standard deviations reported in parentheses and best scores in bold.}
\label{tab:knowledge}
\vspace{-4mm}
\end{table*}

We compared our proposed KEA, using ELECTRA as a base language model, against representative baselines described in Section \ref{sec:baselines}. We found that KEA-infused models have improved performance on fine-grained emotion classification for each of the three datasets, spanning diverse types of input text (i.e. tweets, Reddit comments and emotionally-grounded conversation data). This indicates that enhancing contextual representations using KEA helps encode complex emotions. Overall, Sentence-level KEA (KEA-ELECTRA$_\text{sentence}$) performs the best on most of the evaluation metrics. We note that word-level KEA (KEA-ELECTRA$_\text{word}$) only offers marginal improvement over the baseline approaches as compared to sentence-level KEA. In addition, it is interesting to note that word-level knowledge incorporation done by both $\text{k-ELECTRA}_{\text{BiLSTM}}$ and $\text{KEA-ELECTRA}_{\text{word}}$ have decreased performance compared to their sentence-level counterparts $\text{k-ELECTRA}_{\text{concat}}$ and KEA-ELECTRA respectively.

\subsection{Generalizing to other contextual encoders}
To understand whether the proposed method extends to other pre-trained models, we evaluate our approach using BERT \cite{devlin-etal-2019-bert} as the contextual encoder. 
Table \ref{tab:context} shows that word-level KEA does not exhibit improved performance over BERT. On the other hand, KEA-BERT$_\text{sentence}$ outperforms BERT on all the datasets on almost all of the metrics, indicating that the sentence-level KEA seems to be a more generalizable way of incorporating knowledge into contextual encoders. One possible reason for the low generalizability of word-level knowledge incorporation is that the tasks at hand have sentence-level text inputs. As a result, sentence-level KEA which extracts global information by taking the entire input into consideration encodes the emotional information in the input in a better fashion. However, it will be interesting in the future to understand how word-level knowledge could still be incorporated to bolster local information dependent tasks such as word-level sentiment analysis.

\subsection{Extension to other knowledge sources}
\label{sec:other_knowledge}
To further investigate the versatility of KEA in integrating external knowledge into pre-trained language models, we made use of another lexicon, NRC Emotion Intensity Lexicon (NRC-EIL) \cite{mohammad2018word} (henceforth referred to as EIL). This lexicon data provides intensity annotations which contains real-valued scores of intensity based on eight basic emotions (anger, anticipation, disgust, fear, joy, sadness, surprise, and trust) for 10k English words. In this analysis, we compare ELECTRA which does not have emotional knowledge against the three knowledge-incorporating models, $\text{k-ELECTRA}_{\text{concat}}$ which concatenates the knowledge embedding at the end, $\text{KEA-ELECTRA}_{\text{word}}$ which adds knowledge in the word-level and $\text{KEA-ELECTRA}_{\text{sentence}}$ which adds knowledge in the sentence-level. 


We can see from Table \ref{tab:knowledge} that $\text{KEA-ELECTRA}_\text{word}$ only marginally improves the performance over ELECTRA whereas $\text{KEA-ELECTRA}_{\text{sentence}}$ outperforms ELECTRA and $\text{KEA-ELECTRA}_\text{word}$, incorporating emotional knowledge more effectively to help recognise the complex set of emotions. Interestingly, while the direct concatenation of contextual representation and emotional knowledge (i.e. $\text{k-ELECTRA}_{\text{concat}}$) showed improved performance when compared to ELECTRA in the case of VAD from Table \ref{tab:main_res}, incorporating EIL instead decreased the performance by a large margin. This highlights that incorporating knowledge in the right manner is key to preserve the rich information learnt by the pre-trained models, and we can see that employing KEA shows consistent performance across knowledge sources. 

\subsection{Case Study: Fine-grained Emotion Recognition }
In this case study and error analysis, we use the EmpatheticDialogues dataset to delve deeper into the model behaviour. We classified the majority of misclassifications into two categories, \textbf{C1}: \textit{Emotions with differing intensities} such as \{annoyed,  angry, furious\}, \{afraid, terrified\} and \{joyful, excited\} and, \textbf{C2}: \textit{Emotions with nuanced differences} such as \{nostalgic, sentimental\}, \{embarrassed, ashamed\} and \{impressed, proud\}. 
These label sets are challenging because the emotions within each set are similar, making it difficult for the model to distinguish them. 
In addition, inter-speaker differences in how they identify emotions \cite{kotti2012speaker} could increase the difficulty. Table \ref{speaker-relative} depicts two similar conversations, where Speaker \emph{A} labelled feeling \textit{terrified} while Speaker \emph{B} labelled feeling \textit{afraid} (despite actually using the word ``terrified'').

\begin{table}[h]
\centering
\resizebox{0.38\textwidth}{!}{%
\begin{tabular}{l|ll}
\hline
Speaker & Conversation snippet & Label    \\ \hline
\textbf{A} & \begin{tabular}[c]{@{}l@{}} ``I  am  so  scared  to  live  in  my  \\ neighborhood ... There are people \\ that come around shooting their\\ guns.....'' \end{tabular}              & Terrified \\ 
\textbf{B} & \begin{tabular}[c]{@{}l@{}} ``I was terrified to walk home from \\ the bar one night ... There were \\ gunshots nearby so I just ran home \\ as fast as I could ...'' \end{tabular} & Afraid
\end{tabular}}
\caption{Two examples from the EmpatheticDialogues dataset depicting the variation observed amongst speakers for similar contexts.}
\label{speaker-relative}
\vspace{-5mm}
\end{table}
\begin{figure}[h!]
\centering
    \resizebox{7cm}{!}{\includegraphics[width=\linewidth,trim=0 1cm 0 0]{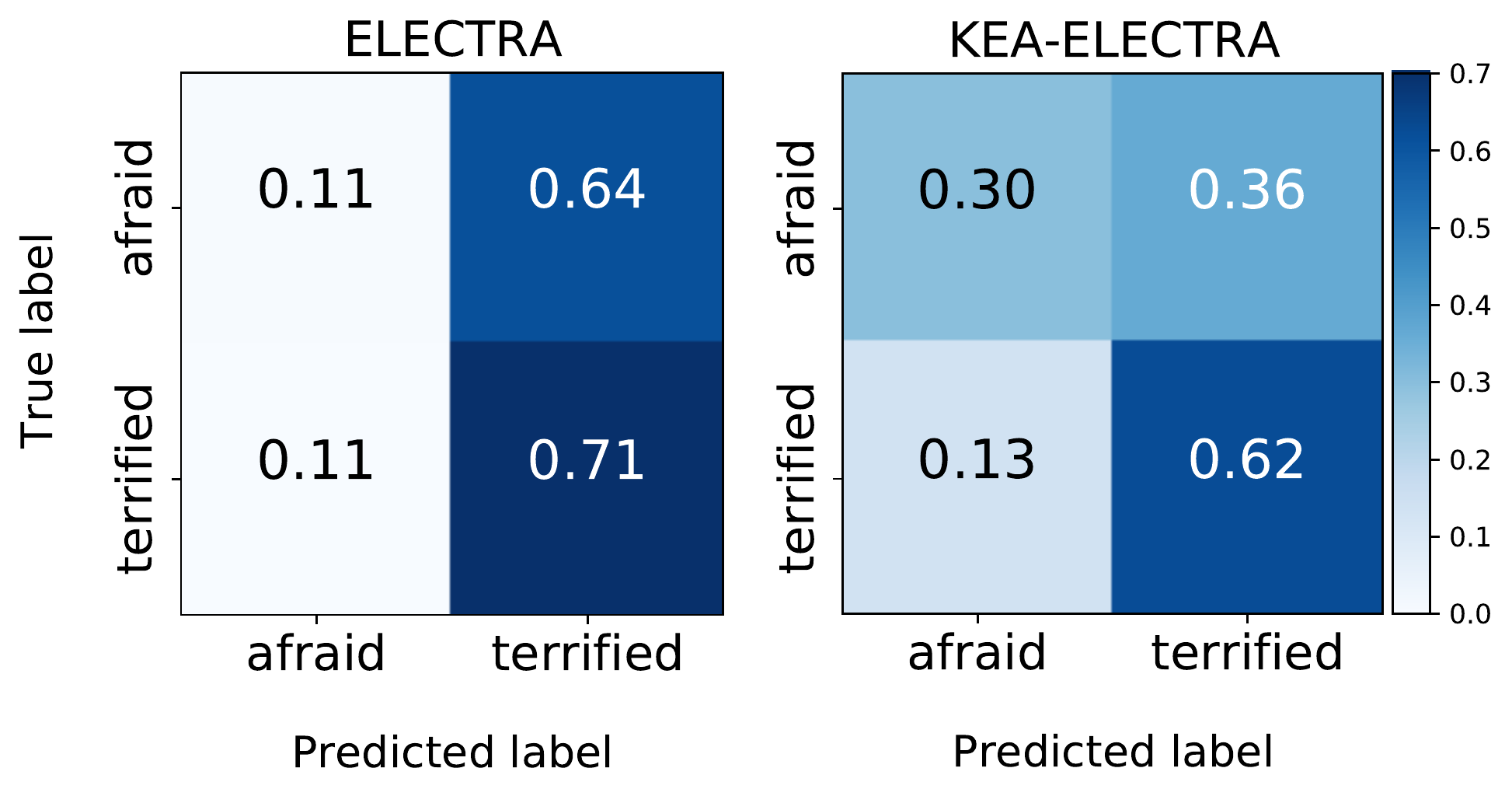}} \\
    \resizebox{7cm}{!}{\includegraphics[width=\linewidth]{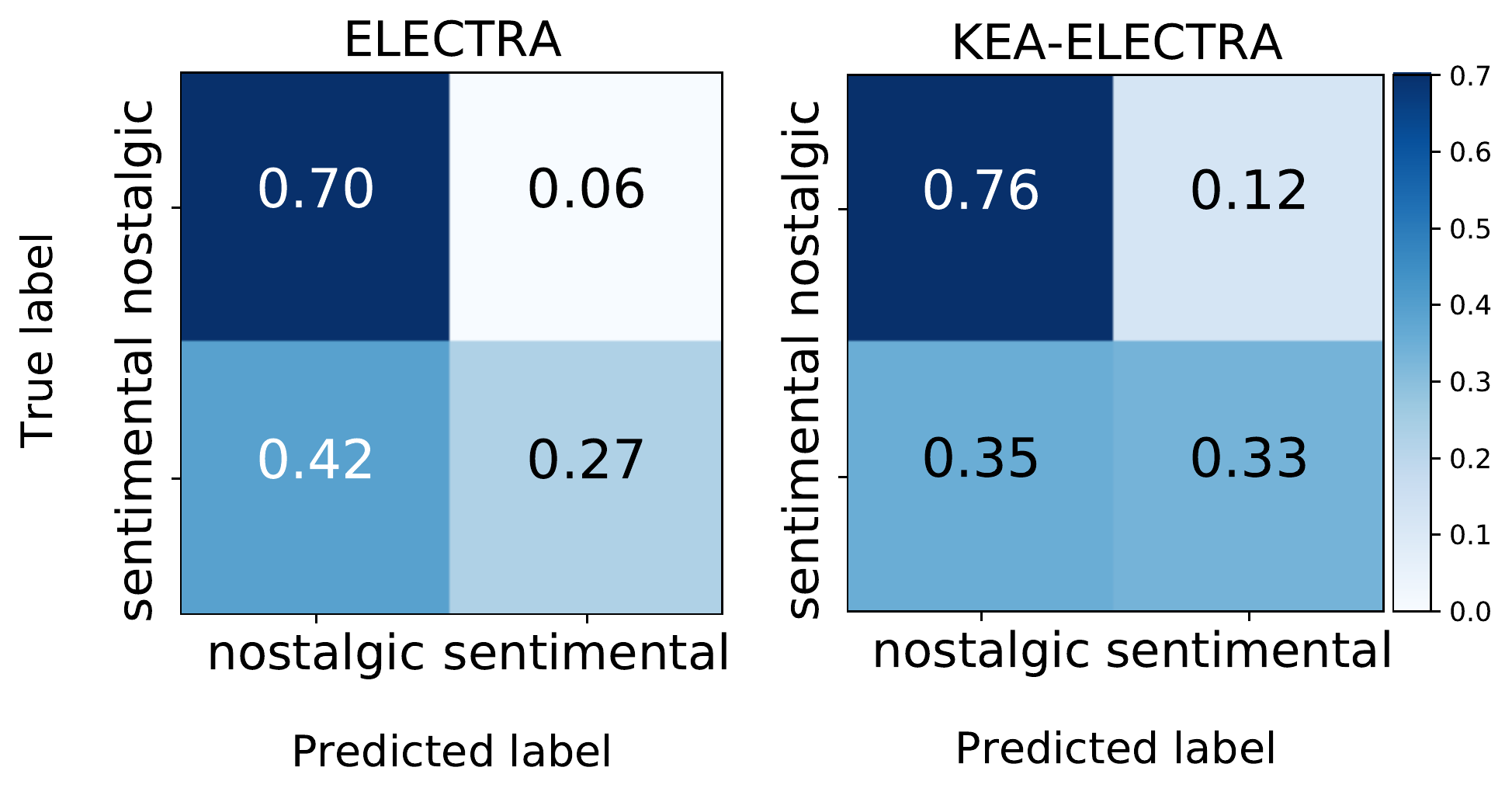}}
  \caption{Excerpts from the confusion matrices to show comparisons between ELECTRA and $\text{KEA-ELECTRA}_{\text{sentence}}$ for the two categories of misclassifications. Top row: \textit{(afraid, terrified)}. ELECTRA tends to classify both as \textit{terrified}, while $\text{KEA-ELECTRA}_{\text{sentence}}$ shows a marked improvement in classifying \textit{afraid}, though at the cost of some correct classification of \textit{terrified}.
  Bottom row: \textit{(nostalgic, sentimental)}. The KEA improvement here is  marginal.}
  \label{fig:small_cm} 
\end{figure}

Next, we turn to how KEA improved performance, by comparing ELECTRA and $\text{KEA-ELECTRA}_{\text{sentence}}$ in Figure \ref{fig:small_cm}. 
We show snippets from the confusion matrix from both models to compare misclassifications amongst problematic label sets. In the example for C1 \{\textit{afraid, terrified}\}, KEA-based model fares better for both the emotion classes by reducing misclassification amongst them. For category C2 \{\textit{nostalgic, sentimental}\}, there is not much improvement; this can be attributed to the interchangeable usage of these emotion labels in conversational language. We have provided more  comparisons and entire confusion matrices in the Supplementary material.
\subsection{Limitations and Future Work}
\label{sec:limitations}

Although incorporating external emotional knowledge via KEA to pre-trained language models improves fine-grained emotion recognition, there are a number of outstanding challenges. First, the fine-grained nature of the emotion classes has not been explicitly encoded into our model architecture. Developing such model architectures that inherently capture the subtleties amongst the emotion classes could help create better representations, which could lead to improved emotion recognition performance. Second, the inter-individual variability in that exists while expressing emotions is a limitation, as seen in Table \ref{speaker-relative} where they use different labels for similar contexts. Modelling this variability is a highly challenging task. A potential solution to this problem could involve actively fine-tuning emotion recognition models to specific users. Third, while we have shown the efficacy of our model using two knowledge sources, these sources are similar in nature---in our case they both are emotion lexicons, where the ``knowledge" is represented using real-valued numbers. Future work could explore how KEA can be extended to incorporate differing types of knowledge sources such as knowledge graphs, categorical data, and relational knowledge \cite{roy2020incorporating} and also delve deeper into the effect that the kind of knowledge has in recognising different types of emotions. Another promising direction is the use of external knowledge to help in few-shot learning scenarios. The knowledge sources have information regarding emotion labels that do not belong to the current task which can potentially help an existing model trained with external knowledge learn unseen labels using fewer data samples. This work is a start towards equipping deep-learning models to recognise larger number of emotions and in the future we aim to address the above-mentioned challenges. 

\subsection{Ethics Statement}

Finally, we want to end on a note about ethical affective computing. At a broader level, emotion recognition technology has been coming under increasing scrutiny, due to two sets of factors: (i) increasing awareness of the limitations of technology to accurately ``understand'' human emotions (e.g., see \cite{barrett2019emotional}, for limitations with facial expressions), and (ii) the deployment of such technology in applications that directly impact people \cite{crawford2019ai}. Our work does not speak to (ii), but it does directly address (i), in that the motivation of our paper includes increasing the scope of text emotion classification models to go beyond six emotions. As we highlighted in the introduction, AI models today are trained on too few emotions, and this severely limits the scientific validity of these models, as well as limiting confidence in their deployment in real-life scenarios. We hope that our work, sustained over time and together with other researchers in the field, would strengthen the confidence people have about the validity of such emotion recognition technology. This will be part of an ongoing conversation to improve our technology and alleviate some of the concerns surrounding their development and deployment \cite{ong2021ethical}. 

\section{Conclusion}
In this work, we propose using KEA (Knowledge Embedded Attention) to incorporate emotional knowledge from external knowledge sources (like emotion lexicons) into contextual representations provided by pre-trained language models like ELECTRA. 
Across our various analyses with different contextual encoders (BERT) and other knowledge sources, we find that our Sentence-Level KEA tends to perform very well across the three datasets we considered (Tweets, Reddit posts, and online conversations), and reduces misclassification of several commonly confusable sets of emotions.
This work provides a strong example of how we can improve our AI to be more emotionally-intelligent. If we want our AI to be sensitive and know when to offer condolences versus when to play an upbeat song, we need to be mindful to train our AI to handle more complex and fine-grained emotions, and at the same time modelling and being sensitive to psychological nuances, like inter-individual variation in emotion experience and expressions \cite{ong2019modeling, siegel2018emotion, cowen2017self}.

\section*{Acknowledgment}
This research was done on publically available datasets.
\bibliographystyle{IEEEtran}
\bibliography{biblio}

\begin{thebibliography}{10}
\providecommand{\url}[1]{#1}
\csname url@samestyle\endcsname
\providecommand{\newblock}{\relax}
\providecommand{\bibinfo}[2]{#2}
\providecommand{\BIBentrySTDinterwordspacing}{\spaceskip=0pt\relax}
\providecommand{\BIBentryALTinterwordstretchfactor}{4}
\providecommand{\BIBentryALTinterwordspacing}{\spaceskip=\fontdimen2\font plus
\BIBentryALTinterwordstretchfactor\fontdimen3\font minus
  \fontdimen4\font\relax}
\providecommand{\BIBforeignlanguage}[2]{{%
\expandafter\ifx\csname l@#1\endcsname\relax
\typeout{** WARNING: IEEEtran.bst: No hyphenation pattern has been}%
\typeout{** loaded for the language `#1'. Using the pattern for}%
\typeout{** the default language instead.}%
\else
\language=\csname l@#1\endcsname
\fi
#2}}
\providecommand{\BIBdecl}{\relax}
\BIBdecl

\bibitem{huang2020challenges}
M.~Huang, X.~Zhu, and J.~Gao, ``Challenges in building intelligent open-domain
  dialog systems,'' \emph{ACM Trans. Inf. Syst.}, vol.~38, no.~3, Apr. 2020.

\bibitem{poria2019emotion}
S.~Poria, N.~Majumder, R.~Mihalcea, and E.~Hovy, ``Emotion recognition in
  conversation: Research challenges, datasets, and recent advances,''
  \emph{IEEE Access}, vol.~7, pp. 100\,943--100\,953, 2019.

\bibitem{alswaidan2020survey}
N.~Alswaidan and M.~E.~B. Menai, ``A survey of state-of-the-art approaches for
  emotion recognition in text,'' \emph{Knowl. Inf. Syst.}, vol.~62, no.~8, pp.
  2937--2987, Aug. 2020.

\bibitem{Skerry15neuralrepresentations}
A.~E. Skerry, R.~Saxe, A.~E. Skerry, and R.~Saxe, ``Neural representations of
  emotion are organized around abstract event features,'' \emph{Curr. Biology},
  vol.~25, no.~15, pp. 1945--54, 2015.

\bibitem{cowen2017self}
A.~S. Cowen and D.~Keltner, ``Self-report captures 27 distinct categories of
  emotion bridged by continuous gradients,'' \emph{Proceedings of the National
  Academy of Sciences}, vol. 114, no.~38, pp. E7900--E7909, 2017.

\bibitem{scherer2013human}
K.~R. Scherer and B.~Meuleman, ``Human emotion experiences can be predicted on
  theoretical grounds: Evidence from verbal labeling,'' \emph{PLOS ONE},
  vol.~8, no.~3, pp. 1--8, 03 2013.

\bibitem{rashkin-etal-2019-towards}
H.~Rashkin, E.~M. Smith, M.~Li, and Y.-L. Boureau, ``Towards empathetic
  open-domain conversation models: A new benchmark and dataset,'' in
  \emph{Proceedings of the 57th Annual Meeting of the Association for
  Computational Linguistics}, Jul. 2019, pp. 5370--5381.

\bibitem{demszky-etal-2020-goemotions}
D.~Demszky, D.~Movshovitz-Attias, J.~Ko, A.~Cowen, G.~Nemade, and S.~Ravi,
  ``{G}o{E}motions: A dataset of fine-grained emotions,'' in \emph{Proceedings
  of the 58th Annual Meeting of the Association for Computational Linguistics},
  Jul. 2020.

\bibitem{clark2020electra}
K.~Clark, M.-T. Luong, Q.~V. Le, and C.~D. Manning, ``{ELECTRA}: Pre-training
  text encoders as discriminators rather than generators,'' in
  \emph{International Conference on Learning Representations}, 2020.

\bibitem{devlin-etal-2019-bert}
J.~Devlin, M.-W. Chang, K.~Lee, and K.~Toutanova, ``{BERT}: Pre-training of
  deep bidirectional transformers for language understanding,'' in
  \emph{Proceedings of the 2019 Conference of the North {A}merican Chapter of
  the Association for Computational Linguistics: Human Language Technologies},
  Jun. 2019, pp. 4171--4186.

\bibitem{roy2020incorporating}
A.~Roy and S.~Pan, ``Incorporating extra knowledge to enhance word embedding,''
  in \emph{Proceedings of the 29th International Joint Conference on Artificial
  Intelligence, {IJCAI-20}}, 2020, pp. 4929--4935.

\bibitem{shin2017lexicon}
B.~Shin, T.~Lee, and J.~D. Choi, ``Lexicon integrated {CNN} models with
  attention for sentiment analysis,'' in \emph{Proceedings of the 8th Workshop
  on Computational Approaches to Subjectivity, Sentiment and Social Media
  Analysis}, 2017, pp. 149--158.

\bibitem{su2018lstm}
M.~{Su}, C.~{Wu}, K.~{Huang}, and Q.~{Hong}, ``Lstm-based text emotion
  recognition using semantic and emotional word vectors,'' in \emph{2018 First
  Asian Conference on Affective Comput. and Intell. Interaction (ACII Asia)},
  2018, pp. 1--6.

\bibitem{colneric2020emotion}
N.~{Colnerič} and J.~{Demšar}, ``Emotion recognition on twitter: Comparative
  study and training a unison model,'' \emph{IEEE Transactions on Affective
  Computing}, vol.~11, no.~3, pp. 433--446, 2020.

\bibitem{wu2019attending}
Z.~Wu, X.~Zhang, T.~Zhi-Xuan, J.~Zaki, and D.~C. Ong, ``Attending to emotional
  narratives,'' in \emph{2019 8th International Conference on Affective
  Computing and Intelligent Interaction (ACII)}, 2019, pp. 648--654.

\bibitem{huang2019emotionx}
Y.-H. Huang, S.-R. Lee, M.-Y. Ma, Y.-H. Chen, Y.-W. Yu, and Y.-S. Chen,
  ``Emotionx-idea: Emotion bert--an affectional model for conversation,''
  \emph{arXiv:1908.06264}, 2019.

\bibitem{ekman1999basic}
P.~Ekman, \emph{Basic Emotions}.\hskip 1em plus 0.5em minus 0.4em\relax John
  Wiley \& Sons, Ltd, 1999.

\bibitem{robert2001the}
R.~Plutchik, ``The nature of emotions,'' \emph{American Scientist}, vol.~89,
  no.~4, pp. 344--350, 2001.

\bibitem{de2019joint}
L.~De~Bruyne, P.~Atanasova, and I.~Augenstein, ``Joint emotion label space
  modelling for affect lexica,'' \emph{arXiv:1911.08782}, 2019.

\bibitem{margatina-etal-2019-attention}
K.~Margatina, C.~Baziotis, and A.~Potamianos, ``Attention-based conditioning
  methods for external knowledge integration,'' in \emph{Proceedings of the
  57th Annual Meeting of the Association for Computational Linguistics}, Jul.
  2019, pp. 3944--3951.

\bibitem{ma2018sentic}
Y.~Ma, H.~Peng, T.~Khan, E.~Cambria, and A.~Hussain, ``Sentic lstm: a hybrid
  network for targeted aspect-based sentiment analysis,'' \emph{Cognitive
  Computation}, vol.~10, no.~4, pp. 639--650, 2018.

\bibitem{poerner-etal-2020-e}
N.~Poerner, U.~Waltinger, and H.~Sch{\"u}tze, ``{E}-{BERT}:
  Efficient-yet-effective entity embeddings for {BERT},'' in \emph{Findings of
  the Association for Computational Linguistics: EMNLP 2020}, Nov. 2020, pp.
  803--818.

\bibitem{zhang-etal-2019-ernie}
Z.~Zhang, X.~Han, Z.~Liu, X.~Jiang, M.~Sun, and Q.~Liu, ``{ERNIE}: Enhanced
  language representation with informative entities,'' in \emph{Proceedings of
  the 57th Annual Meeting of the Association for Computational Linguistics},
  Jul. 2019, pp. 1441--1451.

\bibitem{levine-etal-2020-sensebert}
Y.~Levine, B.~Lenz, O.~Dagan, O.~Ram, D.~Padnos, O.~Sharir, S.~Shalev-Shwartz,
  A.~Shashua, and Y.~Shoham, ``{S}ense{BERT}: Driving some sense into {BERT},''
  in \emph{Proceedings of the 58th Annual Meeting of the Association for
  Computational Linguistics}, Jul. 2020, pp. 4656--4667.

\bibitem{tian-etal-2020-skep}
H.~Tian, C.~Gao, X.~Xiao, H.~Liu, B.~He, H.~Wu, H.~Wang, and F.~Wu, ``{SKEP}:
  Sentiment knowledge enhanced pre-training for sentiment analysis,'' in
  \emph{Proceedings of the 58th Annual Meeting of the Association for
  Computational Linguistics}, Jul. 2020, pp. 4067--4076.

\bibitem{zhong2019knowledge}
P.~Zhong, D.~Wang, and C.~Miao, ``Knowledge-enriched transformer for emotion
  detection in textual conversations,'' in \emph{Proceedings of the 2019
  Conference on Empirical Methods in NLP and the 9th Intl. Joint Conference on
  NLP (EMNLP-IJCNLP)}, 2019, pp. 165--176.

\bibitem{ke-etal-2020-sentilare}
P.~Ke, H.~Ji, S.~Liu, X.~Zhu, and M.~Huang, ``{S}enti{LARE}: Sentiment-aware
  language representation learning with linguistic knowledge,'' in
  \emph{Proceedings of the 2020 Conference on Empirical Methods in NLP
  (EMNLP)}, Nov. 2020, pp. 6975--6988.

\bibitem{chen2020kgpt}
\BIBentryALTinterwordspacing
W.~Chen, Y.~Su, X.~Yan, and W.~Y. Wang, ``{KGPT}: Knowledge-grounded
  pre-training for data-to-text generation,'' in \emph{Proceedings of the 2020
  Conference on Empirical Methods in Natural Language Processing
  (EMNLP)}.\hskip 1em plus 0.5em minus 0.4em\relax Association for
  Computational Linguistics, 2020, pp. 8635--8648. [Online]. Available:
  \url{https://aclanthology.org/2020.emnlp-main.697}
\BIBentrySTDinterwordspacing

\bibitem{liu2021kg}
Y.~Liu, Y.~Wan, L.~He, H.~Peng, and S.~Y. Philip, ``{KG-BART}: Knowledge
  graph-augmented {BART} for generative commonsense reasoning,'' in
  \emph{Proceedings of the AAAI Conference on Artificial Intelligence},
  vol.~35, no.~7, 2021, pp. 6418--6425.

\bibitem{wu2020contextguided}
Z.~Wu and D.~C. Ong, ``Context-guided {BERT} for targeted aspect-based
  sentiment analysis,'' in \emph{Proceedings of the 35th AAAI Conference on
  Artificial Intelligence}, 2021.

\bibitem{babanejad-etal-2020-affective}
N.~Babanejad, H.~Davoudi, A.~An, and M.~Papagelis, ``Affective and contextual
  embedding for sarcasm detection,'' in \emph{Proceedings of the 28th
  International Conference on Computational Linguistics}, Dec. 2020, pp.
  225--243.

\bibitem{ostendorff2019enriching}
M.~Ostendorff, P.~Bourgonje, M.~Berger, J.~Moreno-Schneider, G.~Rehm, and
  B.~Gipp, ``Enriching bert with knowledge graph embeddings for document
  classification,'' \emph{arXiv:1909.08402}, 2019.

\bibitem{wang2020kadapter}
R.~Wang, D.~Tang, N.~Duan, Z.~Wei, X.~Huang, J.~ji, G.~Cao, D.~Jiang, and
  M.~Zhou, ``{K-Adapter}: Infusing knowledge into pre-trained models with
  adapters,'' \emph{arXiv e-prints}, 2020.

\bibitem{mohammad2018word}
S.~M. Mohammad, ``Word affect intensities,'' in \emph{Proceedings of the 11th
  Ed. of the Language Resources and Evaluation Conference}, 2018.

\bibitem{mohammad-2018-obtaining}
S.~Mohammad, ``Obtaining reliable human ratings of valence, arousal, and
  dominance for 20,000 {E}nglish words,'' in \emph{Proceedings of the 56th
  Annual Meeting of the Association for Computational Linguistics}, Melbourne,
  Australia, Jul. 2018, pp. 174--184.

\bibitem{vaswani2017attention}
A.~Vaswani, N.~Shazeer, N.~Parmar, J.~Uszkoreit, L.~Jones, A.~N. Gomez, L.~u.
  Kaiser, and I.~Polosukhin, ``Attention is all you need,'' in \emph{Advances
  in Neural Inf. Processing Sys.}, I.~Guyon, U.~V. Luxburg, S.~Bengio,
  H.~Wallach, R.~Fergus, S.~Vishwanathan, and R.~Garnett, Eds., vol.~30, 2017,
  pp. 5998--6008.

\bibitem{mohammad-kiritchenko-2018-understanding}
S.~Mohammad and S.~Kiritchenko, ``Understanding emotions: A dataset of tweets
  to study interactions between affect categories,'' in \emph{Proceedings of
  the 11th International Conference on Language Resources and Evaluation}, May
  2018.

\bibitem{lin2017structured}
Z.~Lin, M.~Feng, C.~N.~d. Santos, M.~Yu, B.~Xiang, B.~Zhou, and Y.~Bengio, ``A
  structured self-attentive sentence embedding,'' \emph{arXiv:1703.03130},
  2017.

\bibitem{poria-etal-2017-context}
S.~Poria, E.~Cambria, D.~Hazarika, N.~Majumder, A.~Zadeh, and L.-P. Morency,
  ``Context-dependent sentiment analysis in user-generated videos,'' in
  \emph{Proceedings of the 55th Annual Meeting of the Association for
  Computational Linguistics}, 2017, pp. 873--883.

\bibitem{lai2015recurrent}
S.~Lai, L.~Xu, K.~Liu, and J.~Zhao, ``Recurrent convolutional neural networks
  for text classification,'' \emph{AAAI Conference on Artificial Intelligence},
  2015.

\bibitem{speer2017conceptnet}
R.~Speer, J.~Chin, and C.~Havasi, ``Conceptnet 5.5: An open multilingual graph
  of general knowledge,'' in \emph{Proceedings of the 31st AAAI Conference on
  Artificial Intelligence}, 2017, p. 4444–4451.

\bibitem{jiao2020real}
W.~Jiao, M.~Lyu, and I.~King, ``Real-time emotion recognition via attention
  gated hierarchical memory network,'' in \emph{Proceedings of the AAAI
  Conference on Artificial Intelligence}, 2020, pp. 8002--8009.

\bibitem{baziotis2018ntua}
C.~Baziotis, A.~Nikolaos, A.~Chronopoulou, A.~Kolovou, G.~Paraskevopoulos,
  N.~Ellinas, S.~Narayanan, and A.~Potamianos, ``{NTUA-SLP} at {SemEval}-2018
  task 1: Predicting affective content in tweets with deep attentive {RNNs} and
  transfer learning,'' in \emph{Proceedings of The 12th International Workshop
  on Semantic Evaluation}, 2018, pp. 245--255.

\bibitem{kingma2015Adam}
D.~P. Kingma and J.~Ba, ``Adam: {A} method for stochastic optimization,'' in
  \emph{3rd International Conference on Learning Representations}, 2015.

\bibitem{kotti2012speaker}
M.~Kotti and F.~Patern{\`o}, ``Speaker-independent emotion recognition
  exploiting a psychologically-inspired binary cascade classification schema,''
  \emph{International Journal of Speech Technology}, pp. 131--150, 2012.

\bibitem{barrett2019emotional}
L.~F. Barrett, R.~Adolphs, S.~Marsella, A.~M. Martinez, and S.~D. Pollak,
  ``Emotional expressions reconsidered: Challenges to inferring emotion from
  human facial movements,'' \emph{Psychological Science in the Public
  Interest}, vol.~20, no.~1, pp. 1--68, 2019.

\bibitem{crawford2019ai}
K.~Crawford, R.~Dobbe, T.~Dryer, G.~Fried, B.~Green, E.~Kaziunas, A.~Kak,
  V.~Mathur, E.~McElroy, A.~N. S{\'a}nchez \emph{et~al.}, ``{AI Now} 2019
  report,'' \emph{New York, NY: AI Now Institute}, 2019.

\bibitem{ong2021ethical}
D.~C. Ong, ``An ethical framework for guiding the development of
  affectively-aware artificial intelligence,'' in \emph{2021 9th International
  Conference on Affective Computing and Intelligent Interaction (ACII)}, 2021.

\bibitem{ong2019modeling}
D.~Ong, Z.~Wu, Z.-X. Tan, M.~Reddan, I.~Kahhale, A.~Mattek, and J.~Zaki,
  ``Modeling emotion in complex stories: the {S}tanford {E}motional
  {N}arratives {D}ataset,'' \emph{IEEE Trans. on Affective Comput.}, 2019.

\bibitem{siegel2018emotion}
E.~H. Siegel, M.~K. Sands, W.~Van~den Noortgate, P.~Condon, Y.~Chang, J.~Dy,
  K.~S. Quigley, and L.~F. Barrett, ``Emotion fingerprints or emotion
  populations? {A} meta-analytic investigation of autonomic features of emotion
  categories,'' \emph{Psychological Bulletin}, vol. 144, no.~4, p. 343—393,
  2018.

\end{thebibliography}

\end{document}